# EMPIRICAL COMPARISON OF AGENT COMMUNICATION PROTOCOLS FOR TASK ORCHESTRATION

**Dobrovolskyi I. V.** – MS in AI & ML Engineering, Independent Researcher, Sunnyvale, CA, USA.
ORCID: https://orcid.org/0009-0005-6938-0384.

## ABSTRACT

**Context.** The problem of comparative evaluation of communication protocols for task orchestration by large language model (LLM) agents is considered. The object of study is the process of interaction between LLM agents and external tools, as well as between autonomous LLM agents, during task orchestration.

**Objective.** The goal of this work is to develop a systematic pilot benchmark comparing tool integration, multi-agent delegation, and hybrid architectures for standardized queries at three levels of complexity, and to quantify the advantages and disadvantages in terms of response time, context window consumption, cost, error recovery, and implementation complexity.

**Method.** An open-source health analyzer application was developed to query four public data sources for repository statistics, package downloads, vulnerability data, and community metrics. Thirty queries were run five times for each architecture at three levels of complexity, resulting in 450 fail-safe executions. The same large language model was used in all executions. The statistical analysis used bootstrap confidence intervals, the Mann–Whitney test with Bonferroni correction, and Cliff's delta effect sizes.

**Results.** A statistically significant crossover effect was found: the tool integration protocol is faster for simple queries, but the multi-agent protocol outperforms on complex multi-project comparisons, consuming fewer tokens through distributed context windows, resulting in lower cost. The hybrid architecture tracks the winner at each level of complexity with near-optimal performance.

**Conclusions.** The scientific novelty is an empirical demonstration that the two protocols exhibit a complexity-dependent performance crossover, showing that they complement each other rather than compete. Context-window bloat in single-agent architectures is quantified for the first time in a pilot case study. The practical significance is a validated decision framework: using a tool integration protocol for single-source queries, a multi-agent protocol for complex orchestration, and a hybrid routing when complexity is unknown at runtime. Directions for further research include extending the benchmark to additional models, distributed deployments, and additional application domains.

## ABBREVIATIONS

A2A is an Agent-to-Agent Protocol;
API is an Application Programming Interface;
AST is an Abstract Syntax Tree;
CI is a Confidence Interval;
HTTP is a Hypertext Transfer Protocol;
JSON-RPC is a JSON Remote Procedure Call;
LLM is a Large Language Model;
LOC is a line of Code;
MCP is a Model Context Protocol;
SDK is a Software Development Kit;
stdio is a Standard Input/Output;

## NOMENCLATURE

$A$ is a set of architectural configurations;
$c$ is a complexity tier;
$f$ is a weighted objective function;
$n$ is a number of executions per cell;
$P$ is a performance vector (latency, tokens, cost);
$p$ is a statistical significance (p-value);
$Q$ is a set of benchmark queries;
$\delta$ is a Cliff's delta effect size.

## INTRODUCTION

The rapid development of large language model (LLM) agent systems has raised a fundamental architectural question: how should agents interact with tools and with each other? Between 2025 and 2026, two protocols have become de facto standards.

The Model Context Protocol (MCP), introduced by Anthropic [1], standardizes how LLM agents can call external LLM tools. An agent connects to MCP servers via stdio or HTTP transport, checks for available tools, and calls them within a single context window. The MCP SDK has exceeded 97 million downloads monthly [2].

The Agent-to-Agent Protocol (A2A), introduced by Google [3], allows for autonomous agents to discover, interact with, and delegate tasks to other agents over HTTP using JSON-RPC. Each agent publishes an Agent Card at /.well-known/agent.json to check capabilities. The A2A has been launched with support by more than 50 technology partners [4].

The most relevant prior work is a survey by Ehtesham et al. [5], which treats MCP, ACP, A2A, and ANP as complementary layers in the agent interaction stack but provides no empirical measurements.

**The object of study** is the communication process between LLM agents and external tools or other agents during task orchestration.

**The subject of study** is the comparative performance characteristics of MCP, A2A, and Hybrid communication architectures at different levels of query complexity.

**The purpose of the work** is to develop the empirical benchmark for comparing agent communication protocols and to define a decision-making framework for protocol selection. To achieve this goal, the following tasks were completed: developing three equivalent architectures (MCP-only, multi-agent A2A, Hybrid) implementing the same application logic; defining 30 standardized queries at three complexity levels; conducting 450 controlled experiments and collecting six performance metrics; applying nonparametric statistical tests to quantify the significance and magnitude of the observed differences; and developing a practical protocol selection framework based on query complexity.

## 1 PROBLEM STATEMENT

Given a set of natural language queries $Q = \{q_1, q_2, \ldots, q_{30}\}$ classified into three complexity levels (simple, medium, complex), and three architectural configurations $A \in \{\text{MCP, A2A, Hybrid}\}$, the objective is to determine for each complexity level which architecture minimizes a composite performance vector $P$ = (latency, tokens, cost) subject to correctness constraints.

Formally, for each complexity level $c \in$ {simple, medium, complex}, the goal is to identify $A^* = \text{argmin}_a f(P_a,c)$, where $f$ is a weighted objective function. The constraints require zero execution failures (reliability = 1.0) and zero hallucinations in the output.

The input variables are 30 standardized queries and four public API data sources (GitHub, npm, OSV.dev, Stack Overflow). The output variables are six metrics: wall-clock latency, total number of LLM tokens, cost in USD, number of LLM API calls, number of data source API calls, and code complexity. The quality assessment criteria are statistical significance (Mann-Whitney U test) and effect size magnitude (Cliff's delta).

## 2 REVIEW OF THE LITERATURE

The tool augmentation of LLMs was standardized by Schick et al. [6] with Toolformer, which demonstrated that language models can learn to invoke external tools during text generation. Patil et al. [7] extended this with Gorilla, enabling API-level tool invocation with improved accuracy. Yao et al. [8] introduced ReAct, which demonstrated the reasoning-action loop that underpins modern agent frameworks. The work mentioned above establishes tool use capabilities but does not address inter-agent communication protocols.

Multi-agent LLM orchestration has been explored through several frameworks. Wu et al. [9] proposed AutoGen for conversable multi-agent systems. Moura [10] introduced CrewAI for role-based agent orchestration. Hong et al. [11] developed MetaGPT for collaborative software engineering through multi-agent coordination. All three frameworks implement their own proprietary communication mechanisms but do not use standardized protocols such as MCP or A2A.

MCP was released by Anthropic in November 2024 [1] as an open standard for tool integration. Google introduced A2A in April 2025 [3] as an open protocol for inter-agent delegation. Ehtesham et al. [5] surveyed four agent-interoperability protocols (MCP, ACP, A2A, ANP) and proposed a phased-adoption roadmap that treats MCP and A2A as complementary rather than competing. Their analysis is qualitative and does not include empirical measurements. No previous study has benchmarked these protocols head-to-head with controlled experiments. The unsolved part of the problem is the absence of empirical data to guide practitioners in selecting between MCP and A2A for specific use cases. The present work addresses this gap.

## 3 MATERIALS AND METHODS

An Open Source Health Analyzer was built as the experimental application. It answers natural-language queries about open-source project health by pulling data from four public APIs: the GitHub REST API for repository statistics, the npm Registry for package downloads, OSV.dev for vulnerability data, and Stack Overflow for community metrics. The same application logic was implemented in three architectures described below.

In Architecture A (MCP-Only), a single LLM agent connects to four MCP tool servers via stdio transport. The agent decides which tools to call, executes them within its tool-use loop, and accumulates all tool responses in a single context window (Fig. 1).

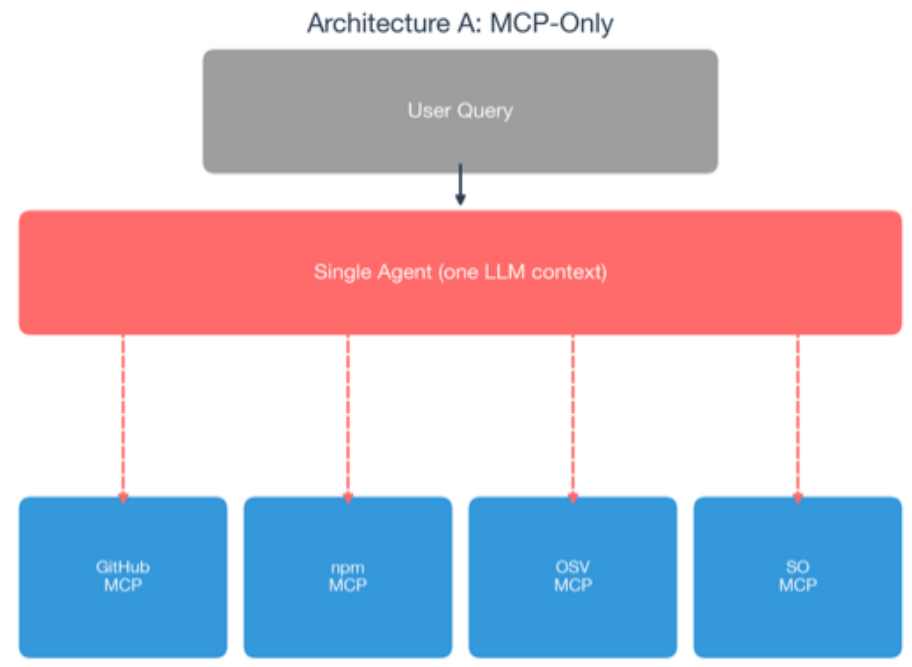

Figure 1 – MCP-Only architecture: single agent with four stdio tool servers

In Architecture B (A2A Multi-Agent), a coordinator agent delegates subtasks to four specialist agents over HTTP using the A2A protocol. Each specialist runs as an independent FastAPI server, publishes an Agent Card at /.well-known/agent.json, and processes tasks via JSON-RPC. The coordinator discovers agents, plans a delegation, dispatches them in parallel, and synthesizes the results. Each specialist preserves its own LLM context window (Fig. 2).

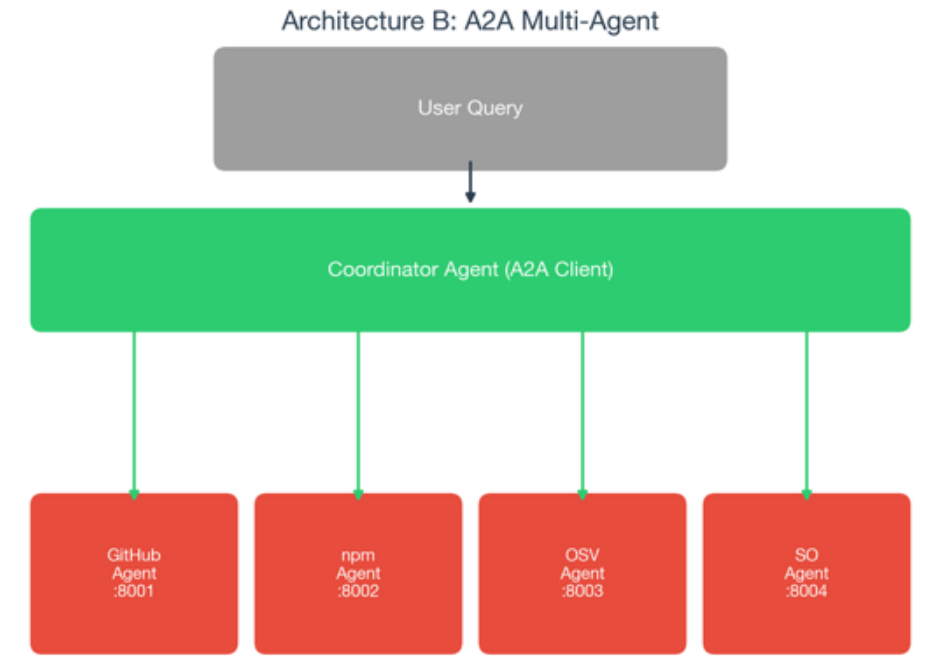

Figure 2 – A2A Multi-Agent architecture: coordinator delegates to four independent HTTP agent servers

In Architecture C (Hybrid), a routing layer classifies query complexity using keyword-based heuristics at zero LLM cost. Simple queries are routed directly through MCP; complex queries are delegated via A2A (Fig. 3).

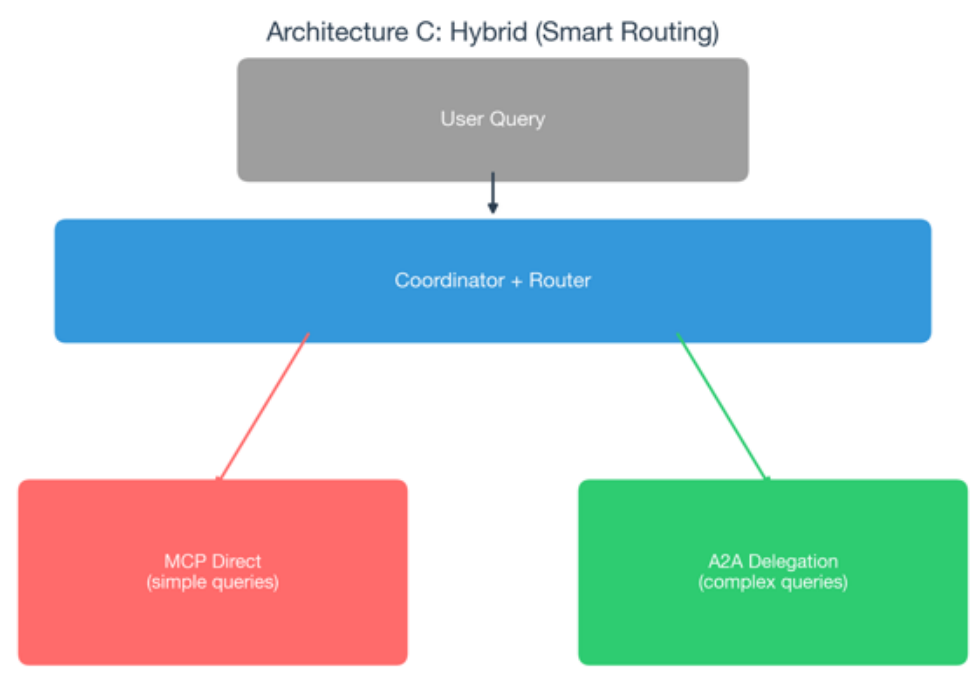

Figure 3 – Hybrid architecture: heuristic router selects protocol per query

Thirty queries were defined across three complexity levels: ten simple queries (single-source, single-project), ten medium queries (multi-source, single-project), and ten complex queries (multi-source, multi-project comparison). Each query was executed five times per architecture, yielding a total of 450 executions. All executions used Claude Sonnet (claude-sonnet-4-20250514) to control for model variation.

Six metrics were recorded for each execution: wall-clock latency in milliseconds, total LLM tokens (prompt plus completion), cost in USD at the rate of \$3 per million input tokens and \$15 per million output tokens, number of LLM API calls, number of data source API calls, and code complexity measured in lines of code and cyclomatic complexity via AST analysis.

For statistical analysis, 95% bootstrap confidence intervals were computed using 10,000 resamples for all means. Pairwise comparisons used the Mann–Whitney U test [12] with Bonferroni correction for nine simultaneous tests. Effect sizes were computed via Cliff's delta [13] and interpreted using the thresholds proposed by Romano et al. [14]: negligible ($|\delta| < 0.147$), small ($0.147 \le |\delta| < 0.33$), medium ($0.33 \le |\delta| < 0.474$), and large ($|\delta| \ge 0.474$).

## 4 EXPERIMENTS

All experiments were conducted on a single machine with all A2A agents deployed on localhost to eliminate network variability. Each of the 30 queries was executed five times per architecture (MCP, A2A, Hybrid), producing 450 total executions. The four data source APIs (GitHub, npm, OSV.dev, Stack Overflow) were queried live during each run; no caching was employed.

Numerical claims in the output (download counts, star counts, version numbers, CVE IDs) were compared programmatically against the JSON values returned by the corresponding APIs. APIs. No divergences were detected. All 450 executions completed successfully with zero failures.

## 5 RESULTS

Table 1 shows the results summary across all 450 executions.

Table 1 – Summary results across 450 executions

| Complexity | Architecture | Latency, s | Tokens | Cost, \$ | n |
|---|---|---|---|---|---|
| Simple | MCP | 8.9 | 4,671 | 0.0169 | 50 |
| Simple | A2A | 19.8 | 1,988 | 0.0162 | 50 |
| Simple | Hybrid | 10.6 | 4,040 | 0.0162 | 50 |
| Medium | MCP | 30.3 | 9,965 | 0.0435 | 50 |
| Medium | A2A | 36.0 | 6,420 | 0.0521 | 50 |
| Medium | Hybrid | 34.8 | 6,556 | 0.0476 | 50 |
| Complex | MCP | 51.8 | 34,959 | 0.1297 | 50 |
| Complex | A2A | 45.1 | 11,318 | 0.0789 | 50 |
| Complex | Hybrid | 45.9 | 16,421 | 0.0941 | 50 |

Figure 4 shows the latency crossover, which is the central finding. For simple queries, MCP (8.9 s) is much faster than A2A (19.8 s) with $p < 0.0001$ and a large effect ($\delta = -0.90$). For medium queries, MCP (30.3 s) retains its advantage ($p = 0.0001$, $\delta = -0.51$). For complex queries, A2A (45.1 s) outperforms MCP (51.8 s) with a small-to-medium effect ($\delta = +0.29$). The Hybrid architecture tracks the winner at each complexity level.

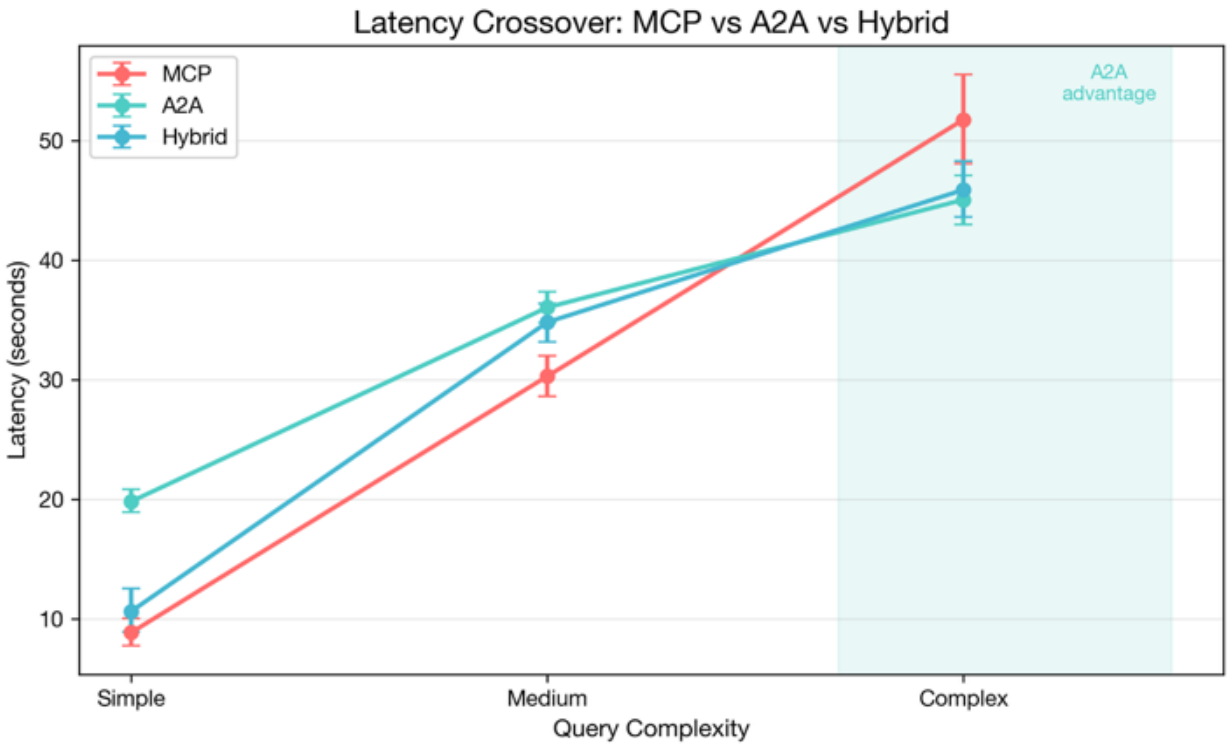

Figure 4 – Latency crossover: MCP wins on simple and medium queries; A2A wins on complex queries (95% bootstrap CI shown as error bars)

Token consumption explains the crossover. For complex queries, MCP consumes 34,959 tokens, which is 3.1 times more than A2A (11,318 tokens; $p < 0.0001$, $\delta = +0.84$). In the MCP single-agent architecture, every tool response is added to a single growing context window. For complex queries that requiring many API calls, this accumulation becomes the dominant cost driver. A2A distributes context across specialist agents, keeping each agent's context lean.

Cost follows token usage. All three architectures cost approximately \$0.017 for simple queries. For complex queries, MCP costs \$0.130 while A2A costs \$0.079, representing a 39% reduction. One thousand complex queries would cost $130 with MCP versus $79 with A2A at enterprise scale.

All pairwise comparisons at the simple and complex tiers show large effects ($|\delta| \geq 0.474$), confirming that the observed performance differences are not artifacts of sampling variability.

Regarding code complexity, MCP requires 706 LOC across 11 files. A2A requires 1,530 LOC across 15 files, which is 2.2 times more. The Hybrid architecture requires 1,899 LOC. The A2A overhead comes from HTTP server infrastructure (291 LOC) and per-agent boilerplate. Average cyclomatic complexity is 7.9 for MCP and 10.7 for A2A.

## 6 DISCUSSION

Based on the findings, the following decision framework is proposed (Fig. 5). For queries involving a single data source or simple lookups, MCP should be used due to its zero-overhead advantage: no HTTP, no agent discovery, no delegation LLM call. For complex queries involving multiple projects and multiple data sources, A2A should be used because it distributes context across specialist agents, keeping each at approximately 2,263 tokens versus 34,959 tokens in MCP's single window. When query complexity is unknown at runtime, the Hybrid architecture with a lightweight heuristic router achieves near-optimal performance across tiers. The exception is the medium tier, where the A2A uses fewer total tokens than the MCP but has slightly higher monetary cost due to a larger share of output tokens in the multi-agent protocol.

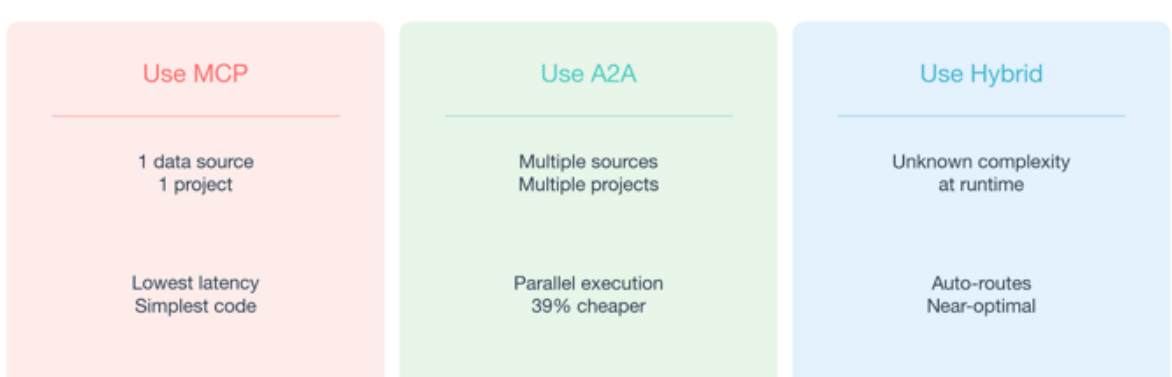

Figure 5 – Decision framework for protocol selection based on query complexity

Two opposing forces drive the crossover. MCP's advantage on simple queries comes from zero overhead: no HTTP connections, no agent discovery, and no delegation LLM call. A2A's advantage on complex queries comes from context isolation: MCP accumulates all tokens in a single window, whereas A2A distributes them across independent agents.

These results imply that MCP and A2A are complementary rather than competing. MCP excels as a tool integration layer (agent-to-tool), while A2A excels as an orchestration layer (agent-to-agent). The A2A specialist agents in this study use direct API calls internally; in production, they could use MCP for their tool connections, combining both protocols.

The present work confirms empirically the complementarity hypothesized qualitatively by Ehtesham et al. [5] and further identifies the specific complexity regime in which the crossover occurs, driven by context-window accumulation.

Several limitations should be noted. All experiments use a single LLM model (Claude Sonnet); the results may not be generalizable to other models. The A2A agents run on a localhost; actual network latency will increase A2A overhead and could shift the crossover point. Only one application domain was tested (open-source projects analysis). Five runs per query (50 per cell) provide good statistical power for detecting large effects but limit the detection of small effects. Different system queries may affect relative performance.

## CONCLUSIONS

A pressing scientific problem of empirically comparing agent communication protocols, previously considered only theoretically, has now been solved.

**The scientific novelty** of these results lies in the fact that a statistically significant, complexity-dependent crossover effect between MCP and A2A has been empirically demonstrated: MCP is faster for simple queries ($p < 0.0001$), while A2A consumes 3.1 times fewer tokens for complex queries, reducing the cost by 39%. This lets practitioners make evidence-based protocol selection decisions rather than relying on intuition. The context window bloat phenomenon in single-agent architectures, previously discussed only qualitatively, has been further developed: it has been quantified as a 3.1-fold increase in token costs, increasing the cost by 39% for complex queries. This enables architects to predict the cost implications of protocol choice.

**The practical significance** of the obtained results lies in the solution structure, validated by 450 executions: use MCP for single-source queries, A2A for complex orchestration of multiple projects, and hybrid routing for unknown execution complexity. The hybrid architecture validates automatic routing with near-optimal performance at each level of complexity. All code, benchmark data, and analysis tools are available under the MIT license at: https://github.com/IvanDobrovolsky/mcp-vs-a2a-bench.

**Prospects for further research** include extending the benchmark to additional LLM models, testing on real-world distributed deployments with network latency, evaluating additional application domains, and developing adaptive routing that explores the data crossover point.

## ACKNOWLEDGEMENTS

The study was performed without financial support.

## DECLARATIONS

**Conflict of interest**: The author declares that there is no conflict of interest in relation to this research, whether financial, personal, authorship, or otherwise, that could affect the research and its results presented in this paper.

**Authors' contributions**: Ivan Dobrovolskyi: Conceptualization, Methodology, Software, Investigation, Data Curation, Writing – Original Draft, Writing – Review & Editing, Visualization.

**Data availability**: The manuscript has associated data in a data repository available at: https://github.com/IvanDobrovolsky/mcp-vs-a2a-bench.

**Software availability**: The manuscript has associated software in a repository available at: https://github.com/IvanDobrovolsky/mcp-vs-a2a-bench.

**Use of artificial intelligence tools**: The authors confirm that they did not use artificial intelligence technologies in creating the submitted work.

# ЕМПІРИЧНЕ ПОРІВНЯННЯ ПРОТОКОЛІВ КОМУНІКАЦІЇ АГЕНТІВ ДЛЯ ОРКЕСТРАЦІЇ ЗАВДАНЬ

**Добровольський І.В.** – магістр зі штучного інтелекту та машинного навчання, незалежний дослідник, Саннівейл, Каліфорнія, США. ORCID: https://orcid.org/0009-0005-6938-0384.

## АНОТАЦІЯ

**Актуальність.** З розвитком агентних систем на основі великих мовних моделей виникли два конкуруючі протоколи комунікації: протокол інтеграції інструментів та протокол міжагентної делегації. Незважаючи на широке впровадження в індустрії десятками партнерів, емпіричне порівняння цих протоколів відсутнє в літературі.

**Мета роботи** – розробка першого систематичного бенчмарку для порівняння архітектур на основі інтеграції інструментів, міжагентної делегації та гібридної на стандартизованих запитах трьох рівнів складності та кількісна оцінка компромісів за часом відгуку, споживанням токенів, вартістю, відновленням після помилок та складністю реалізації.

**Метод.** Було розроблено застосунок для аналізу стану проєктів з відкритим кодом, що звертається до чотирьох публічних джерел даних. Тридцять запитів трьох рівнів складності виконувалися по п'ять разів на кожну архітектуру, що дало кілька сотень виконань без жодного збою. Усі запуски використовували одну й ту саму мовну модель. У статистичному аналізі використовувалися довірчі інтервали бутстрепу, тест Манна-Вітні з корекцією Бонферроні та розміри дельта-ефекту Кліффа.

**Результати.** Було виявлено статистично значущий ефект кросоверу: протокол інтеграції інструментів швидший для простих запитів, але міжагентний протокол переважає його у складних порівняннях кількох проєктів, споживаючи менше токенів через розподіленні контекстні вікна, що призводить до зниження витрат. Гібридна архітектура відстежує переможця на кожному рівні складності з майже оптимальною продуктивністю.

**Висновки.** Наукова новизна полягає в емпіричній демонстрації того, що два протоколи демонструють залежну від складності продуктивність, показуючи, що вони доповнюють один одного, а не конкурують. Вперше кількісно визначено збільшення контекстного вікна в одноагентних архітектурах. Практичне значення полягає у валідованій структурі прийняття рішень: використання протоколу інтеграції інструментів для запитів з одного джерела, багатоагентного протоколу для складної оркестрації та гібридної маршрутизації, коли складність невідома під час виконання. Напрямки подальших досліджень включають розширення бенчмарку на додаткові моделі, розподілені розгортання та додаткові домени застосувань.

## ЛІТЕРАТУРА